# Efficiency Enhancement of
# Probabilistic Model Building Genetic Algorithms

Kumara Sastry
David E. Goldberg
Martin Pelikan

# Efficiency Enhancement of Probabilistic Model Building Genetic Algorithms

Kumara Sastry
Illinois Genetic Algorithms Laboratory (IlliGAL), and
Department of Material Science & Engineering
University of Illinois at Urbana-Champaign
`ksastry@uiuc.edu`

David E. Goldberg
Illinois Genetic Algorithms Laboratory (IlliGAL), and
Department of General Engineering
University of Illinois at Urbana-Champaign
`deg@uiuc.edu`

Martin Pelikan
Department of Math & Computer Science
University of Missouri, St. Louis
`mpelikan@cs.umsl.edu`

**Abstract**

This paper presents two different efficiency-enhancement techniques for probabilistic model building genetic algorithms. The first technique proposes the use of a mutation operator which performs local search in the sub-solution neighborhood identified through the probabilistic model. The second technique proposes building and using an internal probabilistic model of the fitness along with the probabilistic model of variable interactions. The fitness values of some offspring are estimated using the probabilistic model, thereby avoiding computationally expensive function evaluations. The scalability of the aforementioned techniques are analyzed using facetwise models for convergence time and population sizing. The speed-up obtained by each of the methods is predicted and verified with empirical results. The results show that for additively separable problems the competent mutation operator requires $\mathcal{O}(\sqrt{k}\log m)$—where $k$ is the building-block size, and $m$ is the number of building blocks—less function evaluations than its selectorecombinative counterpart. The results also show that the use of an internal probabilistic fitness model reduces the required number of function evaluations to as low as 1-10% and yields a speed-up of 2–50.

## 1 Introduction

A key challenge in genetic and evolutionary computation (GEC) research is the design of *competent* genetic algorithms (GAs). By *competent* we mean GAs that can solve *hard problems*, quickly, reliably, and accurately, and much progress has been made along these lines (Goldberg, 1999; Goldberg, 2002). In essence, competent GA design takes problems that were intractable with first generation GAs and renders them tractable, oftentimes requiring only a subquadratic number of fitness evaluations. However, for large-scale problems, the task of computing even a subquadratic number of function evaluations can be daunting. This is especially the case if the fitness evaluation is a complex simulation, model, or computation. This places a premium on a variety of *efficiency enhancement techniques*. In essence, while competence leads us from *intractability* to *tractability*, efficiency enhancement takes us from *tractability* to *practicality*.

In this paper, we propose two different efficiency-enhancement techniques for competent genetic algorithms in general, and probabilistic model building genetic algorithms in particular: (1) Using a *competent* mutation operator that performs local search in building-block neighborhoods identified by the probabilistic model of variable interactions (Sastry & Goldberg, 2004b; Sastry & Goldberg, 2004a), and (2) Building and Using an *internal probabilistic* model of fitness instead of the more expensive fitness function (Sastry, Pelikan, & Goldberg, 2004; Pelikan & Sastry, 2004). We develop facetwise models to predict the scalability and speed-up of both efficiency-enhancement techniques. Specifically, we summarize and combine the analysis and results presented elsewhere (Sastry & Goldberg, 2004b; Sastry & Goldberg, 2004a; Sastry, Pelikan, & Goldberg, 2004; Pelikan & Sastry, 2004

This paper is organized as follows. The next section gives a brief introduction to extended compact genetic algorithm (eCGA), followed by a description of a scalable mutation algorithm. We analyze the scalability of the mutation algorithm and the speed-up obtained by using it. Section 4 discusses evaluation relaxation in PMBGAs by building and using an internal probabilistic model of fitness. We analyze the scalability and the speed-up of the evaluation-relaxation scheme. Finally, we outline future research directions followed by conclusions.

## 2  Extended Compact Genetic Algorithm (eCGA)

The extended compact GA proposed by Harik (Harik, 1999), like traditional genetic algorithms, is a selectionist search method, where only a subset of better individuals influence the subsequent generation of candidate solutions. However, like other probabilistic model building genetic algorithms (PMBGAs) (Pelikan, Lobo, & Goldberg, 2002), eCGA replaces the traditional variation operators of genetic algorithms by building a probabilistic model of promising solutions and sampling the model to generate new candidate solutions.

The probabilistic model in eCGA is represented by a class of probability models known as marginal product models (MPMs). MPMs are formed as a product of marginal distributions on a partition of the genes. MPMs also facilitate a direct linkage map with each partition separating tightly linked genes. For example, the following MPM, [1,3][2][4], for a four-bit problem represents that the 1$^{\text{st}}$ and 3$^{\text{rd}}$ genes are linked and 2$^{\text{nd}}$ and 4$^{\text{th}}$ genes are independent.

In eCGA, both the structure and the parameters of the model are searched and optimized to best fit the data. To distinguish between better model instances from worse ones, eCGA uses a minimum description length (MDL) metric as the class-selection metric. In essence, the MDL metric penalizes both complex as well as inaccurate models. The MDL metric used in eCGA is a sum of model complexity, $C_m$, and compressed population complexity, $C_p$. In essence, the model complexity, $C_m$, quantifies the model representation size in terms of number of bits required to store all the marginal probabilities. Let, a given problem of size $\ell$ with binary alphabets, have $m$ partitions with $k_i$ genes in the $i^{\text{th}}$ partition, such that $\sum_{i=1}^{m} k_i = \ell$. Then each partition $i$ requires $2^{k_i} - 1$ independent frequencies to completely define its marginal distribution. Furthermore, each frequency is of size $\log_2(n)$, where $n$ is the population size. Therefore, the model complexity $C_m$,

is given by

$$C_m = \log_2(n) \sum_{i=1}^{m} \left(2^{k_i} - 1\right). \tag{1}$$

The compressed population complexity, $C_p$, represents the cost of using a simple model as against a complex one. In essence, the compressed population complexity, $C_p$, quantifies the data compression in terms of the entropy of the marginal distribution over all partitions. Therefore, $C_p$ is evaluated as

$$C_p = n \sum_{i=1}^{m} \sum_{j=1}^{2^{k_i}} -p_{ij} \log_2 (p_{ij}) \tag{2}$$

where $p_{ij}$ is the frequency of the $j^{\text{th}}$ gene sequence of the genes belonging to the $i^{\text{th}}$ partition. In other words, $p_{ij} = n_{ij}/n$, where $n_{ij}$ is the number of chromosomes in the population (after selection) possessing bit-sequence $j \in [1, 2^{k_i}]$ for $i^{\text{th}}$ partition.

The MDL metric is used to evaluate alternative probabilistic models (chosen from admissible MPMs). Similar to other PMBGAs, eCGA uses a greedy search heuristic is used to find an optimal model of the selected individuals in the population. The greedy-search method begins with models at a low level of complexity (all independent variables), and then adding complexity when it locally improves the MDL metric value. This process continues, until no further improvement is possible.

Once the best probabilistic model is built, the new individuals are created by sampling the probabilistic model. The offspring population are generated by randomly choosing subsets from the current individuals according to the probabilities of the subsets as calculated in the probabilistic model.

The population-sizing and the scalability of PMBGAs in general, and the Bayesian optimization algorithm and eCGA in particular, have been studied elsewhere (Pelikan, Goldberg, & Cantú-Paz, 2000a; Pelikan, Sastry, & Goldberg, 2003; Sastry & Goldberg, 2004a). The models predict that the population size required to solve a problem with $m$ building blocks (BBs) of size $k$ with a failure rate of $\alpha = 1/m$ is given by

$$n \propto 2^k \left(\frac{\sigma_{BB}}{d}\right) m \log m, \tag{3}$$

and the number of function evaluations is given by

$$n_{fe} \propto \left(\frac{\sigma_{BB}}{d}\right) \sqrt{k} \cdot 2^k m^{1.5} \log m, \tag{4}$$

where $\sigma_{BB}$ is fitness-variance of a BB and $d$ is the signal difference between competing BBs (Goldberg, Deb, & Clark, 1992). In other words, the models predict that for additively separable problems, eCGA scales subquadratically, $\mathcal{O}(2^k m^{1.5} \log m)$, with problem size.

## 3  Probabilistic Model Building BB-wise Mutation

As explained in the previous section, eCGA builds marginal product models that yields a direct mapping of linkage groups among successful individuals. The probabilistic model yields a global neighborhood information and can be effectively exploited by a mutation operators that performs local search in the building-block neighborhood. In other words, we can replace a bit-wise mutation method that scales exponentially, by an an *enumerative BB-wise mutation* operator as used elsewhere (Sastry & Goldberg, 2004b), which scales subquadratically with problem size.

The BB-wise mutation uses the best individual, and searches for the best building block for each linkage-group identified by the MPM. For example, if model builder identifies $m$ variable-groups with $k$ variables in each group, the BB-wise algorithm will select the best BB out of $2^k$ possible ones in each of the $m$ partition. Note that the performance of the BB-wise mutation can be slightly improved by using a greedy heuristic to search for the best among competing BBs, however, as shown later, the scalability of the probabilistic model building BB-wise mutation operator is determined by the population-size required to accurately identify the building blocks.

It should be noted that while eCGA can only build linkage groups with non-overlapping variables, the mutation procedure can be easily used with other linkage identification techniques that can handle overlapping BBs such as BOA (Pelikan, Goldberg, & Cantú-Paz, 2000b) or DSMDGA (Yu, Goldberg, Yassine, & Chen, 2003). However, since the effect of overlapping interactions between variables is similar to that of an exogenous noise (Goldberg, 2002), crossover is likely to be more effective than mutation (Sastry & Goldberg, 2004b).

Moreover, we perform linkage identification only once in the initial generation. This offline linkage identification works well on problems with BBs of nearly equal salience. However, for problems with BBs of non-uniform salience, we would have to perform linkage identification and update BB information in regular intervals. Furthermore, it might be more efficient to utilize both BB-wise mutation and eCGA model sampling simultaneously or sequentially along the lines of *hybridization* (Goldberg & Voessner, 1999; Sinha & Goldberg, 2003; Sinha, 2003) and *time-continuation* (Goldberg, 1999; Srivastava, 2002) techniques.

To reiterate, the objective of this paper is to couple linkage identification with a mutation operator that performs local search in the BB neighborhood and to verify its effectiveness in solving boundedly difficult additively separable problems. Moreover, the aforementioned enhancements can be designed on the proposed competent selectomutative GA.

### 3.1 Scalability of the BB-wise Mutation

As mentioned earlier, eCGA without mutation scales as $\mathcal{O}(2^k m^{1.5} \log m)$. Here, we consider the scalability of the selectomutative GA, which depends on two factors: (1) The population size required to build accurate probabilistic models of the linkage groups, and (2) the total number of evaluations performed by the BB-wise mutation operator to find optimum BBs in all the partitions.

Pelikan and Sastry (Pelikan, Sastry, & Goldberg, 2003) observed that to build accurate models the population size has to be scaled as,

$$\mathcal{O}\left(2^k m^{1.05}\right) \leq n \leq \mathcal{O}\left(2^k m^{2.1}\right). \tag{5}$$

Since we perform the model building only once, the total number of function evaluations scales as the population size. That is,

$$\mathcal{O}\left(2^k m^{1.05}\right) \leq n_{\text{fe},1} \leq \mathcal{O}\left(2^k m^{2.1}\right). \tag{6}$$

During BB-wise mutation, we evaluate $2^k - 1$ individuals for determining the best BBs in each of the $m$ partitions. Therefore, the total number of function evaluations used during BB-wise mutation is

$$n_{\text{fe},2} = \left(2^k - 1\right) m = \mathcal{O}\left(2^k m\right). \tag{7}$$

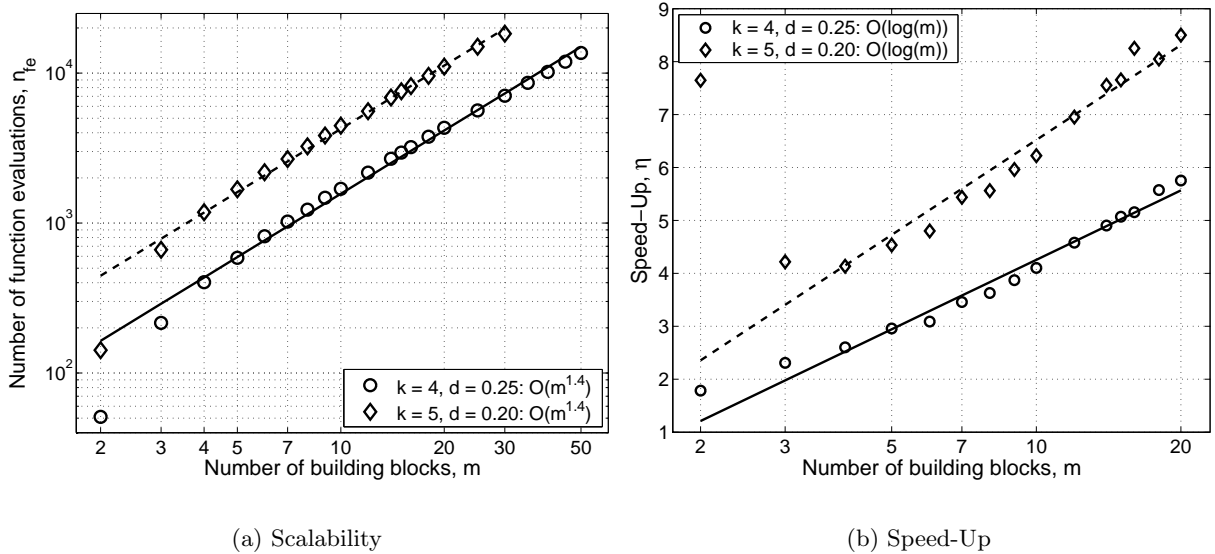

(a) Scalability  (b) Speed-Up

Figure 1: Empirical verification of the scalability and speed-up (Equation 10) obtained by using the probabilistic model building BB-wise mutation over eCGA for the m k-Trap function. The results show that the BB-wise mutation scales as $\mathcal{O}(2^k m^{1.5})$ and the speed-up scales as $\mathcal{O}(\sqrt{k}\log m)$.

From Equations 6 and 7, the total number of function evaluations scales as

$$\mathcal{O}\left(2^k m^{1.05}\right) \leq n_{\text{fe}} \leq \mathcal{O}\left(2^k m^{2.1}\right). \tag{8}$$

Indeed, we empirically observed (Sastry & Goldberg, 2004a) that the number of function evaluations scales as (see Figure 1(a))

$$n_{\text{fe}} = \mathcal{O}(2^k m^{1.5}). \tag{9}$$

The results (Equations 4 and 9) indicate that the selectomutative algorithm is $\mathcal{O}\left(\sqrt{k}\log m\right)$ faster than eCGA in solving boundedly difficult additively separable problems. That is, the speed-up—which is defined as the ratio of number of function evaluations required by eCGA to that required by the selectomutative GA—is given by

$$\eta = \frac{n_{\text{fe}}(\text{eCGA})}{n_{\text{fe}}(\text{BBwise Mutation})} = \mathcal{O}\left(\sqrt{k}\log m\right). \tag{10}$$

Empirical results shown in Figure 1(b) agrees with the above equation. The results show that the probabilistic model building BB-wise mutation is $\mathcal{O}(\sqrt{k}m)$ times faster than the extended compact GA.

## 4 Evaluation-Relaxation in PMBGAs

Evaluation-relaxation schemes are efficiency-enhancement techniques, where an accurate, but computationally-expensive fitness evaluation is replaced (or augmented) with a less-accurate, but computationally less expensive fitness evaluation (Sastry, 2001). The low-cost, less-accurate fitness estimate can either be (1) *Exogenous*, as in the case of surrogate (or, approximate) fitness functions, where an external means is used to develop the fitness estimate (For example, see (Barthelemy & Haftka, 1993)), or (2) *Endogenous*, as in *fitness inheritance* (Smith, Dike, & Stegmann, 1995), where, the fitness estimate is computed internally based on parental fitnesses.

While the use of exogenous models have been extensively—both empirically and analytically—studied (see (Sastry, 2001) and (Jin, 2003) and references therein), limited attention has been paid towards analysis and development of competent methods for building endogenous fitness estimates. The endogenous models used in evolutionary-computation literature are naïve (Smith, Dike, & Stegmann, 1995; Zheng, Julstrom, & Cheng, 1997) and analytically have been shown to yield limited speed-up of about 1.25, both in single-objective (Sastry, Goldberg, & Pelikan, 2001), and in multiobjective cases (Chen, Goldberg, Ho, & Sastry, 2002). In this study, we develop an efficient and effective endogenous probabilistic fitness model, that automatically and adaptively exploits the regularities of the search problem. We show that the speed-up obtained by the use of such a model for estimating the fitness of some offspring, yields significant speed-up.

Similar to earlier studies on fitness inheritance, in the proposed method, all the individuals in the initial population are evaluated and subsequently a portion of the offspring population receives inherited fitness and the other receive actual fitness evaluation. That is, an offspring receives inherited fitness with a probability $p_i$, or an evaluated fitness with a probability $1 - p_i$. However, unlike previous works, which used either average or weighted average of parental fitnesses as the inherited fitness, here we employ the probabilistic model built by eCGA and estimates of linkage-group fitnesses in determining the inherited fitness. Specifically, individuals from parental population that received evaluated fitnesses (that is, individuals whose fitnesses were not estimated) are used to determine the fitnesses of schemata that are defined by the probabilistic model. The schema fitness from different partitions are then used to estimate the fitness of an offspring. The procedure is detailed in the following paragraph.

After the probabilistic model is built and the linkage map is obtained (step 4 of the eCGA algorithm outlined in the previous section), we estimate the fitness of schemata using only those individuals whose fitnesses were not inherited. In all, we estimate the fitness of a total of $\sum_{i=1}^{m} 2^{k_i}$ schemas. Considering our previous example (Section 2) of a four-bit problem, whose model is [1,3][2][4], the schemata whose fitnesses are estimated are: {0*0*, 0*1*, 1*0*, 1*1*, *0**, *1**, ***0, ***1}.

The fitness of a schema, $h$, is defined as the difference between the average fitness of individuals that contain the schema and the average fitness of all the individuals. That is,

$$\hat{f}_s(h) = \frac{1}{n_h} \sum_{\{i|x_i \supset h\}} f(x_i) - \frac{1}{n'} \sum_{i=1}^{n'} f(x_i) \qquad (11)$$

where $n_h$ is the total number of individuals that contain the schema $h$, $x_i$ is the $i^{\text{th}}$ individual and $f(x_i)$ is its fitness, $n'$ is the total number of individuals that were evaluated. If a particular schema is not present in the population, its fitness is arbitrarily set to zero. Furthermore, it should be noted that the above definition of schema fitness is not unique and other estimates can be used. The key point however is the use of the probabilistic model in determining the schema fitnesses.

Once the schema fitnesses across partitions are estimated, the offspring population is created as outlined in Section 2. An offspring receives inherited fitness with a probability $p_i$, referred to as the inheritance probability. The inherited fitness is computed as follows:

$$f_{\text{inh}}(y) = \frac{1}{n'}\sum_{i=1}^{n'} f(x_i) + \sum_{i=1}^{m} \hat{f}_s(h_i \in y) \quad (12)$$

where $y$ is the offspring individual. It should be noted that the eCGA model yields non-overlapping linkage groups and might not be appropriate for problems with overlapping BBs. However, similar concepts can be incorporated in other PMBGAs such as the Bayesian optimization algorithm (BOA) (Pelikan, Goldberg, & Cantú-Paz, 2000b) which can handle overlapping BBs. Moreover, the inherited fitness can be computed in other manner, but the key is to use the estimates of substructure fitnesses in the computation.

With this understanding of the inheritance mechanism, we will now model the effects of fitness inheritance on the scalability of the GA and to predict the speed-up (or efficiency enhancement) obtained through fitness inheritance in the following section.

## 4.1 Scalability of Using the Internal Fitness Model

Elsewhere, facetwise models have been developed for analyzing the effect of using the probabilistic fitness model on the population sizing and convergence time for a GA success (Sastry, Pelikan, & Goldberg, 2004). The population-sizing model is given by

$$n = -c_n \log(\alpha) 2^k \sigma_f^2 (1 + p_i), \quad (13)$$

where $n$ is the population size, $c_n$ is a problem-dependent constant, $k$ is the BB length, $\alpha$ is the probability of failure, and $\sigma_f^2$ is the fitness variance. The convergence-time model is given by

$$t_c = c_t \sqrt{m \cdot k} \sqrt{1 + \frac{\sigma_N^2}{\sigma_f^2}}, \quad (14)$$

where $c_t$ is a problem dependent constant.

We now use the convergence-time and population-sizing models to predict the number of function evaluations required for eCGA success:

$$n_{fe} = n + n(t_c - 1)(1 - p_i). \quad (15)$$

Recall that all the individuals in the initial population are evaluated and there after on an average $n(1 - p_i)$ individuals are evaluated.

To isolate the effect of fitness inheritance on the scalability of eCGA, we consider the ratio of total number of function evaluations required with fitness inheritance and that required without fitness inheritance. That is, we consider the function-evaluation ratio, $n_{fe,r} = n_{fe}/n_{fe}(p_i = 0)$. From Equations 13 and 14, we obtain the following approximation for $n_{fe,r}$:

$$n_{fe,r} \approx (1 + p_i)^{1.5}(1 - p_i). \quad (16)$$

The speed-up that can be obtained through fitness inheritance is given by the inverse of the function-evaluation ratio:

$$\eta_{\text{inh}} = \frac{1}{(1 + p_i)^{1.5}(1 - p_i)}. \quad (17)$$

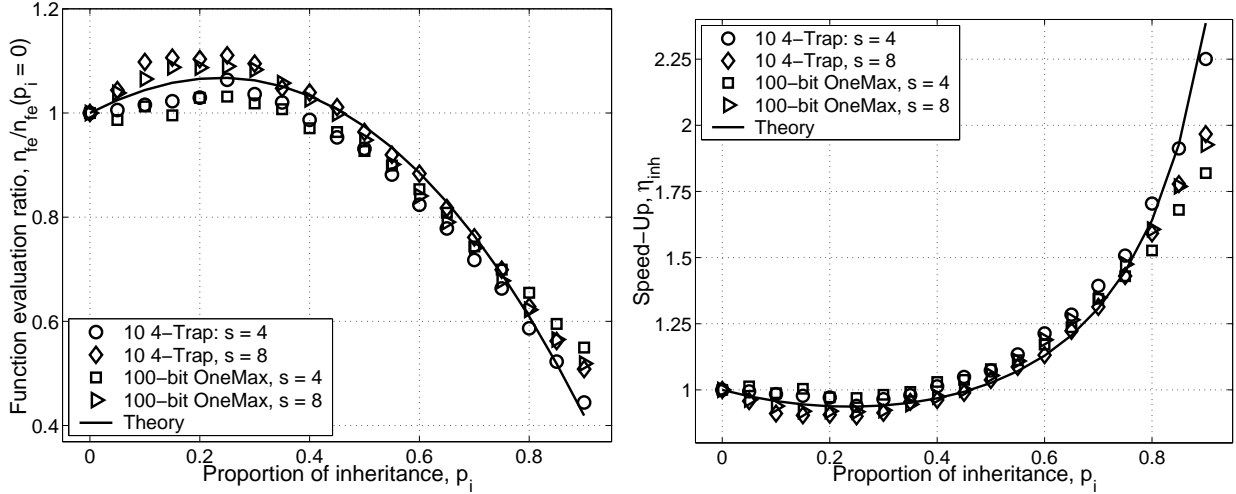

(a) Verification of function-evaluation-ratio model    (b) Verification of speed-up model

Figure 2: Verification of the function-evaluation-ratio model (Equation 16) and the speed-up model (Equation 17) with empirical results on 100-bit OneMax and 10 4-Trap problems. The total number of function evaluations is determined such that the failure probability of an eCGA run is at most $1/m$. The results are averaged over 900–3000 independent runs.

Equation 16 indicates that the function-evaluation ratio increases (or the speed-up reduces) at low $p_i$ values, reaches a maximum at $p_i = 0.2$. When $p_i = 0.2$ the number of function evaluations required is 5% more than that required without inheritance. In other words, the speed-up at $p_i = 0.2$ is 0.95. For inheritance probabilities above 0.2 the function-evaluation ratio decreases (speed-up increases) with the inheritance probability. Equation 17 predicts that the speed-up is maximum when $p_i = 1.0$, however, it should be noted that the models derived are not entirely valid for higher $p_i$ values ($p_i \geq 0.95$). While the fitness model built on eCGA requires fitness evaluation for about 10% of the population, yielding a speed-up of about 1.8–2.25, the fitness model built on BOA requires fitness evaluation for about only 1% of the population, yielding a speed-up of about 35–50 (Pelikan & Sastry, 2004).

The predicted values of function-evaluation-ratio (Equation 16) and the speed-up (Equation 17 are verified with empirical results for OneMax and m k-Trap in Figures 2(a) and 2(b). As shown in Figures 2(a) and 2(b), the empirical results agree with the analytical models. Furthermore, the agreement for the OneMax problem with the models is good even though the building-block identification for the OneMax problem is only partially correct. The results show that the required number of function evaluations is almost of halved with the use of fitness inheritance thereby leading to a speed-up of 1.75–2.25. This is a significant improvement over a speed-up of 1.25 observed for simple GAs with a simple inheritance mechanism. Furthermore, fitness inheritance yields speed-up even when the inheritance probabilities is very high, even as high as 0.85-0.99, which is similar to the empirical observation of Smith, Dike, and Stegmann (Smith, Dike, & Stegmann, 1995). As mentioned earlier, when we used the Bayesian optimization algorithm instead of eCGA, we observed that fitness evaluation was required only for 1% of the population, yielding a speed-up of about 35–50.

Overall, the results suggest that significant efficiency enhancement that can be achieved through an inheritance mechanism that incorporates knowledge of important sub-solutions of a problem and their partial fitnesses.

## 5  Future Work

We demonstrated two efficiency-enhancement techniques for probabilistic model building genetic algorithms that provide significant speed-up. The first method demonstrated the potential of inducting *global* neighborhood information into mutation operations via the automatic discovery of linkage groups by probabilistic model building techniques. The second method illustrated the use of an endogenous probabilistic fitness model for estimating fitness of some individuals instead of the computationally expensive fitness evaluation. The results are very encouraging and warrants further research in the following avenues:

**Hybridization of competent crossover and mutation:** While we considered a bounding case of crossover vs. mutation, it is likely to be more efficient to use an efficient hybrid of competent crossover and mutation operators. For example, we can consider a hybrid GA with *oscillating* populations. A large population is used to gather linkage information and used for crossover, while a small population is used for searching in BB neighborhood.

**Problems with overlapping building blocks:** While this paper considered problems with non-overlapping building blocks, many problems have different building blocks that share common components. The performance of probabilistic model building BB-wise mutation on problems with overlapping building blocks have to be analyzed. Since the effect of overlapping variable interactions is similar to that of exogenous noise (Goldberg, 2002), based on our recent analysis (Sastry & Goldberg, 2004b), a crossover is likely to be more useful than mutation. Moreover, while considering problems with overlapping building blocks, the use of eCGA might not be appropriate, however the fitness model building mechanism should still be valid which can be used in other more sophisticated PMBGAs such as the Bayesian optimization algorithm (Pelikan & Sastry, 2004).

**Problems with non-uniform BB salience:** In this paper we considered additively separable problems with uniform sub-solution salience. Unlike uniformly-scaled problems, in non-uniformly scaled problems BBs are identified sequentially over time. Therefore, in such cases, we would need to regularly update the BB information and develop theory to predict the updating schedule. The effect of non-uniform building-block salience on the speed-up and optimal inheritance proportion should also be investigated.

**Hierarchical problems:** One of the important class of nearly decomposable problems is hierarchical problems, in which the building-block interactions are present at more than a single level. Further investigation is necessary to analyze the performance of BB-wise mutation on hierarchical problems. Additionally, the fitness model building mechanism used in this study could be enhanced and incorporated into hBOA and the efficiency enhancement provided by inheritance can also be investigated.

- **Additional dimensions of problem difficulty:** In this paper we considered one of the dimensions of GA problem difficulty, deception. However, there are other dimensions of problem difficulty (Goldberg, 2002) such as epistasis and external noise. This factors should be included in isolation or in conjunction with other factors of problem difficulty in determining a complete picture of efficiency enhancement provided by fitness inheritance.

- **Real-World problems:** One of the key objectives of analyzing and developing fitness-inheritance mechanism is to aid the principled incorporation of such an mechanism in competent genetic and evolutionary algorithms for successfully solving complex real-world problems in *practical* time.

# 6  Summary & Conclusions

In this paper, we proposed two efficiency-enhancement techniques for probabilistic model building genetic algorithms. The first method is a scalable mutation operator, which performs local search in the neighborhood dictated by the probabilistic model. The second method is an evaluation-relaxation scheme, where an endogenous probabilistic fitness model is developed and used to estimate the fitnesses of some of the offspring instead of expensive function evaluations. We analyze the scalability and speed-up of both the efficiency-enhancement techniques using facetwise models and verify them with empirical results. The results show that for additively-separable problems, the competent mutation operator scales as $\mathcal{O}(2^k m^{1.5})$—where, $k$ is the building-block size, and $m$ is the number of building blocks—and provides a speed-up of $\mathcal{O}(\sqrt{k}\log m)$ over a selectorecombinative PMBGA. The results also show that for additively separable problems, by developing and using an endogenous probabilistic fitness model, only 1–10% of the population requires actual fitness evaluation, providing a speed-up of 2–50.

Overall, the results in this paper demonstrate that the probabilistic model built in PMBGAs can be used in developing a scalable mutation operator and an effective fitness-estimation model, which can both provide significant efficiency enhancement and speed-up the GA process, while yielding high-quality solutions quickly, reliably and accurately. Additionally, since the speed-up provided by the two efficiency-enhancement techniques are nearly independent of each other, the combined speed-up obtained by using both simultaneously should be multiplicative of the individual speed-ups.

## Acknowledgments

This work was sponsored by the Air Force Office of Scientific Research, Air Force Materiel Command, USAF (grant F49620-03-1-0129), the National Science Foundation (ITR grant DMR-99-76550 at MCC, and ITR grant DMR-0121695 at CPSD), and the Dept. of Energy through the Fredrick Seitz MRL (grant DEFG02-91ER45439) at UIUC, and by the Technology Research, Education, and Commercialization Center (TRECC), at UIUC by NCSA and funded by the Office of Naval Research (grant N00014-01-1-0175).